\documentclass[runningheads]{llncs}
\usepackage[T1]{fontenc}
\usepackage{graphicx}
\usepackage{subfig}
\usepackage{multirow}
\usepackage{amsmath}
\usepackage{amssymb}
\usepackage{booktabs}
\usepackage{epsfig}
\usepackage{bbding}
\usepackage{float}
\usepackage{multicol}
\usepackage{diagbox}
\newlength{\colWidth} 
\usepackage{appendix}
\usepackage{algorithm}
\usepackage{algpseudocode}
\usepackage{caption}
\usepackage[colorlinks,
            linkcolor=red,  
            anchorcolor=blue,   
            citecolor=blue,     
            implicit=false,
            ]{hyperref}

\usepackage[table]{xcolor}
\definecolor{mygray}{RGB}{230, 230, 230}
\definecolor{myblue}{RGB}{77, 77, 255}
\definecolor{myred}{RGB}{204, 50, 153}
\definecolor{commentcolor}{RGB}{34, 139, 34} 
\definecolor{codecolor}{RGB}{0, 0, 0} 
\definecolor{algocolor}{RGB}{128, 0, 0} 

\algnewcommand{\LineComment}[1]{\State \textcolor{commentcolor}{\# #1}}

\newcommand{\pseudocodefont}{\fontsize{8pt}{12pt}\selectfont} 

\begin{document}

\bibliographystyle{unsrt}

\title{MambaMIL: Enhancing Long Sequence Modeling with Sequence Reordering in Computational Pathology}

%
%
\titlerunning{MambaMIL}
%

\author{Shu Yang$^\dag$\inst{1} \and
Yihui Wang$^\dag$\inst{1} \and
Hao Chen$^*$\inst{1,2,3}}
%
%
\institute{Department of Computer Science and Engineering, The Hong Kong
University of Science and Technology, Hong Kong, China \and
Department of  Chemical and Biological Engineering, The Hong Kong
University of Science and Technology, Hong Kong, China \and
Division of Life Science, The Hong Kong
University of Science and Technology, Hong Kong, China \\
\email{\{syangcw,ywangrm,jhc\}@cse.ust.hk}}
\maketitle              
\begin{abstract}
Multiple Instance Learning (MIL) has emerged as a dominant paradigm to extract discriminative feature representations within Whole Slide Images (WSIs) in computational pathology. Despite driving notable progress, existing MIL approaches suffer from limitations in facilitating comprehensive and efficient interactions among instances, as well as challenges related to time-consuming computations and overfitting. In this paper, we incorporate the Selective Scan Space State Sequential Model (\textbf{Mamba}) in Multiple Instance Learning (\textbf{MIL}) for long sequence modeling with linear complexity, termed as \textbf{MambaMIL}. By inheriting the capability of vanilla Mamba, MambaMIL demonstrates the ability to comprehensively understand and perceive long sequences of instances. Furthermore, we propose the Sequence Reordering Mamba (SR-Mamba) aware of the order and distribution of instances, which exploits the inherent valuable information embedded within the long sequences. With the SR-Mamba as the core component, MambaMIL can effectively capture more discriminative features and mitigate the challenges associated with overfitting and high computational overhead. Extensive experiments on two public challenging tasks across nine diverse datasets demonstrate that our proposed framework performs favorably against state-of-the-art MIL methods. The code is released at \url{https://github.com/isyangshu/MambaMIL}.

\keywords{Mamba \and  Computational Pathology  \and Whole Slide Images \and Multiple Instance Learning.}
\end{abstract}

\section{Introduction}
\footnote{$^\dag$ means the equal contribution. \\
$^*$ means the corresponding author.}
The digitalization of pathological images into Whole Slide Images (WSIs) has paved the way for computer-aided analysis in computational pathology. However, employing deep learning methods for WSI analysis encounters unique challenges, primarily due to the high resolution of WSIs and the lack of pixel-level annotations. To address these issues, Multiple Instance Learning (MIL)~\cite{amores2013multiple,chen2013multi} has arisen as an ideal solution, where each WSI is represented as a ``bag” and partitioned into a sequence of tissue patches termed ``instances”. With at least one instance being positive, the bag is classified as positive, otherwise negative.

The most widely used paradigm of MIL involves converting instances into low-dimensional features using pre-trained models~\cite{resnet,ctranspath,PLIP}, followed by aggregating these features into bag-level representations for subsequent analysis. 
Under this paradigm, MIL conceptualizes WSI analysis as a long sequence modeling problem, aiming to model the correlation between instances as well as overall contextual information within the entire bag to capture discriminative information. 
Despite the impressive performance, there remain several issues in existing MIL methods. Attention-based methods~\cite{abmil,CLAM,dsmil,dtfdmil} primarily focus on instance-level information based on independent and identical distribution hypotheses. However, these methods neglect the contextual relationships among instances, resulting in inadequate representations of WSIs. Additionally, several methods~\cite{transmil,MCAT,dtmil} utilize Transformer~\cite{vaswani2017attention} for its capability to explore mutual-correlations between instances and model long sequences. Nonetheless, they face significant performance bottlenecks due to extensive computations and overfitting. Overall, the existing methods have limitations in comprehensively mining the contextual information within long sequences, which hinders their performance.

Recently, Structured State Space Sequence (S4)~\cite{s4} has been introduced as an efficient and effective architecture to address the bottleneck concerning long sequence modeling. Furthermore, Selective Scan Space State Sequential Model~\cite{mamba}, namely Mamba, advances S4 in discrete data modeling by employing an input-dependent selection mechanism and a hardware-aware algorithm, which enables Mamba to achieve linear complexity without sacrificing global receptive fields. However, for inherently non-sequential visual data, the direct application of Mamba to a patchified and flattened image would inevitably lead to a constraint in the receptive fields. This limitation stems from the fact that Mamba solely permits interactions between each patch and previously scanned positions, precluding the estimation of relationships with unscanned patches. Unlike typical visual modalities, WSIs contain scattered and scarce positive patches that exhibit weak spatial correlation, which makes them highly suitable for leveraging the robust sequential modeling capabilities of Mamba. Recently, S4MIL~\cite{S4MIL} introduces S4 model to WSI analysis as a multiple instance learner for instance sequences, which demonstrates the effectiveness of SSM in capturing long-range dependencies. Note that it directly adopts the S4 model without fully considering the unique characteristics of WSIs, resulting in sub-optimal results.

Motivated by the above observations, we propose an efficient and effective benchmark MIL model (MambaMIL) with the following contributions: (1) We incorporate the Mamba framework in MIL to tackle the challenges associated with long sequence modeling and overfitting, marking the first application of Mamba in computational pathology. (2) We propose the Sequence Reordering Mamba (SR-Mamba) aware of the order and distribution of instances, which excels at capturing long-range dependencies among scattered positive instances. As the core component of MambaMIL, SR-Mamba is tailored to learn the correlations between instances in both sequential ordering and transpositional ordering, which significantly enhances the capability of the original Mamba to capture more discriminative features. (3) To investigate the effectiveness of MambaMIL, we conduct comprehensive experiments including overall comparison and ablation studies on two challenging tasks across nine datasets, which demonstrates that MambaMIL can achieve superior performance against the state-of-the-art.

\section{Method}
In this section, we start by presenting the preliminaries associated with State Space Models. Subsequently, we elaborate on the MambaMIL and its core component: Sequence Reordering Mamba (SR-Mamba).
\subsection{Preliminaries}
Inspired by State Space Models~\cite{ssm}, structured state space sequence (S4) models have emerged as a promising architecture for effective long sequence modeling. S4 models are defined with four parameters $(\bigtriangleup, A, B, C)$ as linear time-invariant systems, which map stimulation $x(t) \in\mathbb{R}^L$ to response $y(t) \in\mathbb{R}^L$ though an implicit latent state $h(t) \in\mathbb{R}^N$.  The entire progress can be formulated as,
\begin{equation}
        h'(t) = Ah(t) + Bx(t), ~~y(t) = Ch(t),
\label{eq:1}
\end{equation}
where $A\in\mathbb{R}^{N\times N}$ refers to evolution parameter. $B\in\mathbb{R}^{N\times 1}$ and $C\in\mathbb{R}^{N\times 1}$ present projection parameters. S4 models utilize a timescale parameter $\bigtriangleup$ to transform the continuous parameters $A$, $B$ to the discrete parameters $\bar{A}$, $\bar{B}$,
\begin{equation}
        \bar{A} = \exp(\bigtriangleup A),~~\bar{B} = (\bigtriangleup A)^{-1}(\exp(\bigtriangleup A) - I)\cdot \bigtriangleup B.
\label{eq:2}
\end{equation}
After transforming the parameters, we can utilize the discrete parameters to re-frame the Eq.~\ref{eq:1} in the recurrent mode for efficient autoregressive inference,
\begin{equation}
        h_t = \bar{A}h_{t-1} + \bar{B}x_t,~~y_t = Ch_t.
\label{eq:3}
\end{equation}
Alternatively, the models can also compute output through convolutional mode for efficient parallelizable training,
\begin{equation}
        \bar{K} = (C\bar{B}, C\bar{AB}, ..., C\bar{A}^{M-1}\bar{B}),~~y = x * \bar{K}.
\label{eq:4}
\end{equation}
Mamba further integrates selection mechanisms into S4 models to make the parameters be functions of the input with the efficient hardware-aware parallel algorithm. 
Therefore, Mamba can conduct effective and efficient long sequence modeling by selectively propagating or forgetting information along the sequence based on the current token. 

\begin{figure}[t]
    \centering
    \includegraphics[width=0.9\linewidth]{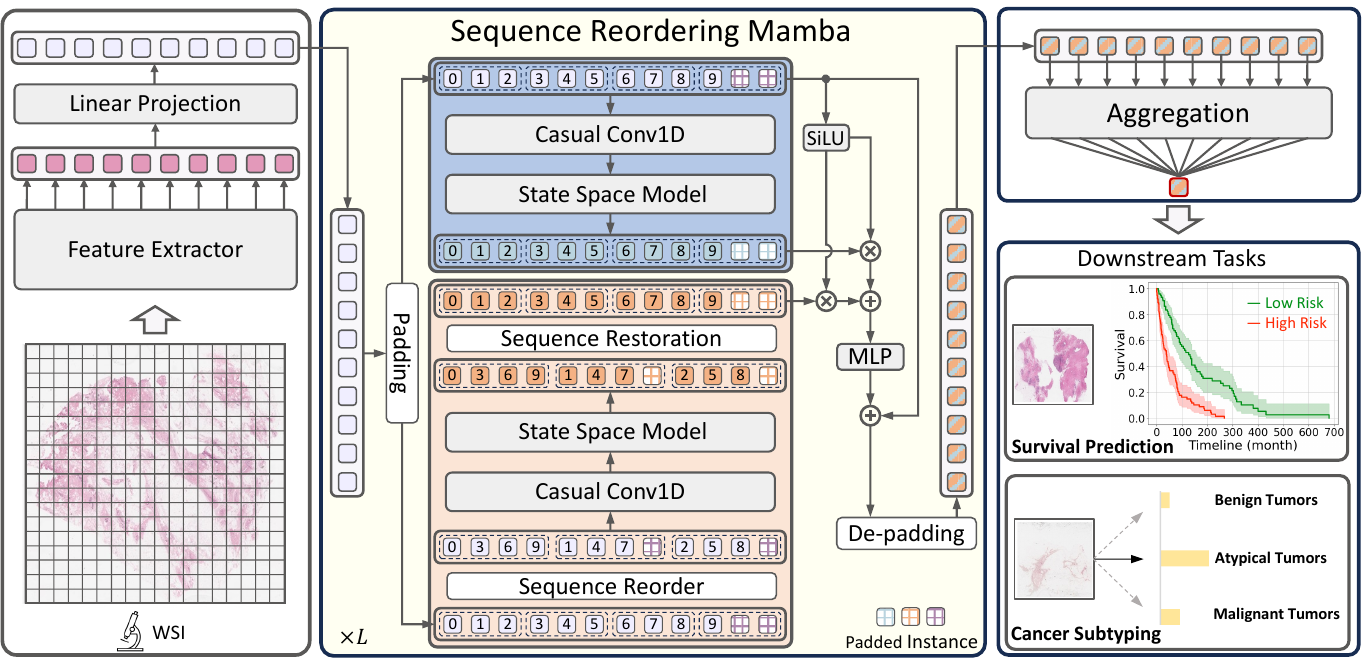}
    \caption{Overview of MambaMIL. Given a set of patches cropped from a slide, we sequentially utilize Feature Extractor, Linear Projection, stacked SR-Mamba modules and Aggregation for WSI analysis.}
    \label{fig:overview}
\end{figure}

\subsection{Overview of MambaMIL}
To efficiently capture the comprehensive contextual information within long sequences of instances, we introduce a novel approach, MambaMIL, by integrating the Mamba framework into MIL, as illustrated in Fig.~\ref{fig:overview}. By inheriting the attributes of Mamba, MambaMIL enables each instance to interact with any of the previously scanned instances through a compressed hidden state, which facilitates effective modeling of long sequences while concurrently mitigating computational complexity.

Specifically, given a WSI, we partition the tissue
regions into a sequence of $L$ patches $\{p_1, p_2,..., p_L\}$, followed by mapping all the patches into instance features $X\in\mathbb{R}^{L\times D}$ by Feature Extractor, where $D$ refers to the feature dimension. Subsequently, the input $X$ is passed through Linear Projection to reduce the dimension. The output is then fed into a series of stacked SR-Mamba modules, which are responsible for modeling long sequences. Finally, we utilize the Aggregation module to obtain bag-level representations for downstream tasks.

\subsection{Sequence Reordering Mamba}
To tackle the restricted receptive fields, we devise the Sequence Reordering Mamba (SR-Mamba) aware of the order and distribution of instances, which exploits the inherent valuable information embedded within the instances. As illustrated in Fig.~\ref{fig:overview}, considering the scattered and scarce positive patches, we establish parallel SSM-based branches upon vanilla Mamba to enhance long sequence modeling. SR-Mamba models two long sequences with distinct sequence orderings, each associated with a unique compressed hidden state, facilitating the learning of more discriminative features.

In detail, given instance features $X\in\mathbb{R}^{L \times D}$, we first partition the sequence of instances into non-overlapping segments of size $R$, and obtain $N=L/R$ segments from the entire sequence. For sequences whose lengths are not divisible by $R$, we pad them with zeros for subsequent reordering. Then the $X$ is fed into two independent branches.
For the first branch, we preserve the original ordering of $X$, which is fed to the subsequent Casual Convolution Layer and State Space Model (SSM) for sequence modeling. The entire process can be formulated as:
\begin{equation}
X' = \operatorname{Norm}(X), \quad Y  = \operatorname{SSM}(\operatorname{SiLU}(\operatorname{Conv1D}(\operatorname{Linear}(X'))).
\end{equation}
Then the $X$ is also used to generate the gating value for $Y$ obtained from SSM, 
\begin{equation}
    Z  = \operatorname{SiLU}(\operatorname{Linear}(X')),\quad X'' = Z\odot Y.
\end{equation}

\begin{figure}[t]
    \centering
    \includegraphics[width=0.95\linewidth]{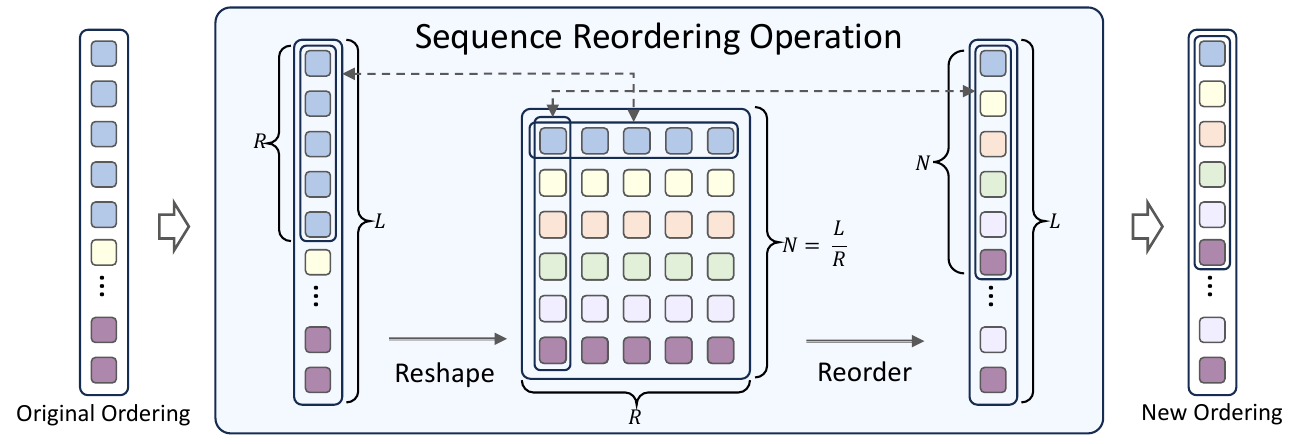}
    \caption{Illustration of Sequence Reordering Operation.}
    \label{fig:SR}
\end{figure}

For the second branch, we propose a Sequence Reordering operation as the core component of SR-Mamba. Specifically, the input instance features are reshaped into a 2-D feature map,  $X\in\mathbb{R}^{L \times D} \xrightarrow{} X_{2d}\in\mathbb{R}^{R \times N \times D}$. We then sample instances from each non-overlapping segment successively along the second dimension of $X_{2d}$, which can be regarded as feature re-embedding. By performing this, we generate the instance features $X_r$ with the new ordering, which can be utilized to embed more discriminative features by the inherent position-sensitive characteristic of Mamba. 
The entire Sequence Reordering operation is depicted in Fig.~\ref{fig:SR}. Then we utilize the subsequent Casual Convolution Layer and State Space Model to model $X_r$, 
\begin{equation}
X_r' = \operatorname{Norm}(X_r'), \quad Y_r  = \operatorname{SSM}(\operatorname{SiLU}(\operatorname{Conv1D}(\operatorname{Linear}(X_r' ))).
\end{equation}
For the enhanced $X_r'$,  we rearrange the sequence into original ordering through partitioning and permutation operations, and gate the instance features by $Z$:
\begin{equation}
    Y_r'  = \psi(Y_r ),\quad X_r'' = Z\odot Y_r',
\end{equation}
where $\psi$ denotes sequence restoration operation. After modeling the long sequences with distinct orderings, we can obtain two discriminative instance features $X''$ and $X_r''$, and aggregate them to obtain $X_{output}$. We devise the aggregation operation as an element-wise addition of the two features,
\begin{equation}
    X_{\text{output}}  = \operatorname{Linear}(X''+X''_r ) + X.
\end{equation}

Distinct from the original Mamba, we maintain the sequential ordering and distribution, while generating new ordering of the instances from a global perspective for feature re-embedding.
Building upon the vanilla Mamba, SR-Mamba is tailored to robustly comprehend and perceive lengthy sequences of instances that are partitioned from WSIs. Built on stacked SR-Mamba modules, MambaMIL is capable of modeling long-range dependencies with linear complexity, resulting in effective model generalization. 

\section{Experiments}
\subsection{Datasets and Evaluation Metrics}
To verify the effectiveness of our proposed MambaMIL, we conduct extensive experiments on two representative downstream tasks across nine public challenging datasets. To investigate generalization and robustness, we utilize two distinct sets of features derived from ResNet-50~\cite{resnet} pre-trained on the ImageNet~\cite{imagenet} and PLIP~\cite{PLIP} pre-trained on 200k pathology image-text pairs. 

\begin{table*}[t]
\centering
\caption{Survival Prediction results on seven main datasets.}
\resizebox{11.2cm}{!}{
\begin{tabular}{c c | c c c c c c c | c}
\toprule
\diagbox{\textbf{Method}}{\textbf{Dataset}} &
& \textbf{BLCA} & \textbf{BRCA} & \textbf{COADREAD} & \textbf{KIRC} & \textbf{KIRP} & \textbf{LUAD} & \textbf{STAD} & \textbf{MEAN}\\
\midrule
\rowcolor{mygray}
\multicolumn{10}{c}{\textbf{\textit{ResNet-50}}}\\
\midrule
Max-Pooling && 0.531$\pm$\scriptsize{0.055} & 0.570$\pm$\scriptsize{0.047} & 0.555$\pm$\scriptsize{0.090} & 0.616$\pm$\scriptsize{0.038} & 0.530$\pm$\scriptsize{0.105} & 0.553$\pm$\scriptsize{0.085} & 0.577$\pm$\scriptsize{0.072} & 0.562\\
Mean-Pooling && 0.595$\pm$\scriptsize{0.067} & 0.602$\pm$\scriptsize{0.057} & 0.592$\pm$\scriptsize{0.109} & 0.660$\pm$\scriptsize{0.039} & 0.691$\pm$\scriptsize{0.073} & 0.602$\pm$\scriptsize{0.045} & 0.595$\pm$\scriptsize{0.059} & 0.620\\
ABMIL~\cite{abmil} && 0.565$\pm$\scriptsize{0.060} & 0.612$\pm$\scriptsize{0.059} & 0.624$\pm$\scriptsize{0.046} & 0.677$\pm$\scriptsize{0.057} & 0.707$\pm$\scriptsize{0.099} & 0.626$\pm$\scriptsize{0.054} & 0.629$\pm$\scriptsize{0.061} & 0.635\\
CLAM-MB~\cite{CLAM} && 0.571$\pm$\scriptsize{0.009} & 0.633$\pm$\scriptsize{0.035} & 0.601$\pm$\scriptsize{0.023} & 0.596$\pm$\scriptsize{0.003} & 0.679$\pm$\scriptsize{0.037} & 0.608$\pm$\scriptsize{0.018} & 0.582$\pm$\scriptsize{0.014} & 0.610\\
DSMIL~\cite{dsmil} && 0.593$\pm$\scriptsize{0.018} & 0.609$\pm$\scriptsize{0.060} & 0.628$\pm$\scriptsize{0.059} & 0.682$\pm$\scriptsize{0.042} & 0.722$\pm$\scriptsize{0.085} & \underline{0.624$\pm$\scriptsize{0.057}} & 0.609$\pm$\scriptsize{0.057} & 0.638\\
DTFDMIL~\cite{dtfdmil} && 0.552$\pm$\scriptsize{0.053} & 0.626$\pm$\scriptsize{0.037} & \underline{0.638$\pm$\scriptsize{0.034}} & 0.687$\pm$\scriptsize{0.075} & 0.724$\pm$\scriptsize{0.102} & 0.623$\pm$\scriptsize{0.048} & 0.619$\pm$\scriptsize{0.073} & 0.638\\
TransMIL~\cite{transmil} && 0.623$\pm$\scriptsize{0.037} & 0.632$\pm$\scriptsize{0.029} & 0.624$\pm$\scriptsize{0.014} & 0.684$\pm$\scriptsize{0.052} & \underline{0.747$\pm$\scriptsize{0.082}} & 0.641$\pm$\scriptsize{0.049} & \underline{0.629$\pm$\scriptsize{0.020}} & \underline{0.654}\\
S4MIL~\cite{S4MIL} && \underline{0.624$\pm$\scriptsize{0.018}} & \underline{0.641$\pm$\scriptsize{0.057}} & 0.608$\pm$\scriptsize{0.049} & \underline{0.691$\pm$\scriptsize{0.039}} & 0.689$\pm$\scriptsize{0.061} & 0.622$\pm$\scriptsize{0.026} & 0.613$\pm$\scriptsize{0.044} & 0.641\\
MambaMIL && \textbf{0.652$\pm$\scriptsize{0.028}} & \textbf{0.675$\pm$\scriptsize{0.065}} & \textbf{0.671$\pm$\scriptsize{0.066}} & \textbf{0.721$\pm$\scriptsize{0.045}} & \textbf{0.748$\pm$\scriptsize{0.094}} & \textbf{0.653$\pm$\scriptsize{0.059}} & \textbf{0.639$\pm$\scriptsize{0.076}} & \textbf{0.680}\\

\midrule
\rowcolor{mygray}
\multicolumn{10}{c}{\textbf{\textit{PLIP}}}\\
\midrule
Max-Pooling && 0.540$\pm$\scriptsize{0.050} & 0.611$\pm$\scriptsize{0.053} & 0.599$\pm$\scriptsize{0.070} & 0.645$\pm$\scriptsize{0.045} & 0.620$\pm$\scriptsize{0.154} & 0.565$\pm$\scriptsize{0.076} & 0.578$\pm$\scriptsize{0.044} & 0.594\\
Mean-Pooling && 0.599$\pm$\scriptsize{0.039} & 0.603$\pm$\scriptsize{0.060} & \underline{0.674$\pm$\scriptsize{0.064}} & 0.669$\pm$\scriptsize{0.065} & 0.766$\pm$\scriptsize{0.063} & 0.617$\pm$\scriptsize{0.048} & 0.603$\pm$\scriptsize{0.052} & 0.647\\
ABMIL~\cite{abmil} && 0.571$\pm$\scriptsize{0.041} & 0.607$\pm$\scriptsize{0.036} & 0.641$\pm$\scriptsize{0.013} & 0.643$\pm$\scriptsize{0.077} & 0.772$\pm$\scriptsize{0.065} & 0.570$\pm$\scriptsize{0.066} & 0.573$\pm$\scriptsize{0.037} & 0.625\\
CLAM-MB~\cite{CLAM} && 0.600$\pm$\scriptsize{0.029} & \underline{0.619$\pm$\scriptsize{0.025}} & 0.628$\pm$\scriptsize{0.031} & 0.597$\pm$\scriptsize{0.022} & 0.722$\pm$\scriptsize{0.063} & 0.603$\pm$\scriptsize{0.026} & 0.593$\pm$\scriptsize{0.020} & 0.623\\
DSMIL~\cite{dsmil} && 0.589$\pm$\scriptsize{0.052} & 0.613$\pm$\scriptsize{0.033} & 0.640$\pm$\scriptsize{0.048} &
0.673$\pm$\scriptsize{0.048} & 0.768$\pm$\scriptsize{0.074} & 0.565$\pm$\scriptsize{0.074} & 0.601$\pm$\scriptsize{0.059} & 0.636\\
DTFDMIL~\cite{dtfdmil} && 0.568$\pm$\scriptsize{0.040} & 0.616$\pm$\scriptsize{0.020} & 0.625$\pm$\scriptsize{0.061} & \underline{0.702$\pm$\scriptsize{0.034}} & 0.772$\pm$\scriptsize{0.096} & 0.624$\pm$\scriptsize{0.032} & 0.624$\pm$\scriptsize{0.032} & 0.647\\
TransMIL~\cite{transmil} && 0.586$\pm$\scriptsize{0.059} & 0.611$\pm$\scriptsize{0.065} & 0.620$\pm$\scriptsize{0.031} & 0.673$\pm$\scriptsize{0.030} & 0.798$\pm$\scriptsize{0.063} & 0.622$\pm$\scriptsize{0.036} & \underline{0.630$\pm$\scriptsize{0.067}} & 0.649\\
S4MIL~\cite{S4MIL} && \underline{0.625$\pm$\scriptsize{0.023}} & 0.614$\pm$\scriptsize{0.051} & 0.657$\pm$\scriptsize{0.065} & 0.695$\pm$\scriptsize{0.026} & \underline{0.799$\pm$\scriptsize{0.055}} & 0.635$\pm$\scriptsize{0.056} & \underline{0.637$\pm$\scriptsize{0.063}} & \underline{0.666}\\
MambaMIL && \textbf{0.677$\pm$\scriptsize{0.053}} & \textbf{0.651$\pm$\scriptsize{0.029}} & \textbf{0.698$\pm$\scriptsize{0.063}} & \textbf{0.715$\pm$\scriptsize{0.049}} & \textbf{0.805$\pm$\scriptsize{0.051}} & \textbf{0.652$\pm$\scriptsize{0.027}} & \textbf{0.653$\pm$\scriptsize{0.253}} & \textbf{0.693}\\
\bottomrule
\end{tabular}}
\label{tab:survival}
\end{table*}

\noindent\textbf{Survival Prediction}. We conduct comprehensive experiments on seven public challenging cancer datasets (\textbf{BLCA}, \textbf{BRCA}, \textbf{COADREAD}, \textbf{KIRC}, \textbf{KIRP}, \textbf{LUAD}, and \textbf{STAD}) from \textbf{TCGA}, containing WSIs annotated with survival outcomes. To reduce the impact of data split on model evaluation, we implement a 5-fold cross-validation approach, partitioning the data into training and validation subsets in a 4:1 ratio. We utilize the cross-validated Concordance Index (C-Index), along with its standard deviation (std), to assess the effectiveness of our proposed MambaMIL.

\noindent\textbf{Cancer Subtyping}. We perform comparative experiments on two public challenging datasets:  \textbf{BRACS}~\cite{bracs} and \textbf{NSCLC}. 
To ensure the robust evaluation of comparison experiments, we employ 10-fold Monte Carlo cross-validation, which partitions the data into training, validation, and testing sets with a ratio of 8:1:1. Additionally, for fair comparisons with existing methods, we also perform experiments on the official split of the BRACS dataset, marked as $\star$ in Table~\ref{tab:subtyping}. Following the standard setting, we adopt the Area Under Curve (AUC) and Accuracy (ACC) metrics along with their standard deviation (std) for evaluation, which provides a reliable assessment less sensitive to class imbalance.

\subsection{Implementation Details}
We present the experimental results of our MambaMIL on nine datasets, in comparison to the following methods: \textbf{(1)} conventional pooling methods, including Mean Pooling and Max Pooling; \textbf{(2)} ABMIL~\cite{abmil} and three distinct variants, including CLAM-MB~\cite{CLAM}, DSMIL~\cite{dsmil} and DTFDMIL~\cite{dtfdmil}; \textbf{(3)} the Transformer-based TransMIL~\cite{transmil}; \textbf{(4)} the SSM-based S4MIL~\cite{S4MIL}. Following common settings, we adopt the same data pre-processing as in the CLAM~\cite{CLAM} and set a learning rate of $2\times10^{-4}$ for these methods to ensure optimal results and enable fair comparisons. In contrast, to mitigate the randomness introduced by atomic operations in the SR-Mamba module during back-propagation, we implement distinct learning rates for training for different datasets. Detailed hyper-parameters can be found in the Appendix. The special adjustment aims to diminish the effect of gradient disparities on convergence, thereby ensuring stability and reproducibility.

\begin{table*}[t]
    \centering
    \caption{Cancer Subtyping results on two main datasets.}
    \resizebox{11.2cm}{!}{
    \begin{tabular}{ c | >{\centering\arraybackslash}p{1.2cm} >{\centering\arraybackslash}p{1.2cm} | >{\centering\arraybackslash}p{\colWidth} >{\centering\arraybackslash}p{\colWidth} | >{\centering\arraybackslash}p{\colWidth} >{\centering\arraybackslash}p{\colWidth} | >{\centering\arraybackslash}p{1.2cm} >{\centering\arraybackslash}p{1.2cm}}
    \toprule
     \multirow{2}{2.8cm}{\diagbox{\textbf{Method}}{\textbf{Dataset}}} &  \multicolumn{2}{c|}{\textbf{BRACS-7$\star$}} & \multicolumn{2}{c|}{\textbf{BRACS-7}} & \multicolumn{2}{c|}{\textbf{NSCLC-2}} & \multicolumn{2}{c}{\textbf{MEAN}}\\
     \cline{2-9}
     & AUC & ACC & AUC & ACC & AUC & ACC & AUC & ACC\\
    \midrule
    \rowcolor{mygray}
    \multicolumn{9}{c}{\textbf{\textit{ResNet-50}}}\\
    \midrule
    Max-Pooling  & 0.630 & 0.241 & 0.707$\pm$\scriptsize{0.053} & 0.389$\pm$\scriptsize{0.066} & 0.943$\pm$\scriptsize{0.019} & 0.869$\pm$\scriptsize{0.017} & 0.760 & 0.500\\
    Mean-Pooling & 0.658 & 0.299 & 0.729$\pm$\scriptsize{0.039} &0.396$\pm$\scriptsize{0.060} & 0.913$\pm$\scriptsize{0.041} & 0.837$\pm$\scriptsize{0.037} & 0.767 & 0.511\\
    ABMIL~\cite{abmil} & 0.715 & 0.230 & 0.765$\pm$\scriptsize{0.041} &0.393$\pm$\scriptsize{0.084} & 0.938$\pm$\scriptsize{0.025} & 0.864$\pm$\scriptsize{0.036} & 0.806 & 0.495 \\
    CLAM-MB~\cite{CLAM} & 0.729 & 0.379 & \underline{0.780$\pm$\scriptsize{0.043}} & \underline{0.457$\pm$\scriptsize{0.073}} & 0.933$\pm$\scriptsize{0.027} & 0.851$\pm$\scriptsize{0.022} & 0.814 & 0.563\\
    DSMIL~\cite{dsmil} & 0.751 & 0.333 & 0.768$\pm$\scriptsize{0.045} & 0.452$\pm$\scriptsize{0.059} & \underline{0.940$\pm$\scriptsize{0.024}} & \underline{0.880$\pm$\scriptsize{0.023}} & \underline{0.820} & 0.555\\
    DTFDMIL~\cite{dtfdmil} & \underline{0.753} & \underline{0.390} & 0.758$\pm$\scriptsize{0.057} &0.448$\pm$\scriptsize{0.049} &  0.928$\pm$\scriptsize{0.055}& 0.835$\pm$\scriptsize{0.031}   
    & 0.813 & \underline{0.558}\\
    TransMIL~\cite{transmil} & 0.613 & 0.310 & 0.699$\pm$\scriptsize{0.040} & 0.363$\pm$\scriptsize{0.073} & 0.937$\pm$\scriptsize{0.019} & 0.846$\pm$\scriptsize{0.044} &0.750 & 0.506\\
    S4MIL & 0.718 & 0.356 & 0.760$\pm$\scriptsize{0.028} & 0.422$\pm$\scriptsize{0.095} & 0.914$\pm$\scriptsize{0.036} & 0.829$\pm$\scriptsize{0.039} &0.797 & 0.536\\
    MambaMIL & \textbf{0.773} & \textbf{0.460} & \textbf{0.804$\pm$\scriptsize{0.028}} & \textbf{0.506$\pm$\scriptsize{0.050}} & \textbf{0.959$\pm$\scriptsize{0.027}} & \textbf{0.891$\pm$\scriptsize{0.044}} & \textbf{0.845} & \textbf{0.619}\\

    \midrule
    \rowcolor{mygray}
    \multicolumn{9}{c}{\textbf{\textit{PLIP}}}\\
    \midrule
    Max-Pooling  & 0.652 & 0.230 & 0.720$\pm$\scriptsize{0.035} & 0.365$\pm$\scriptsize{0.072} & 0.941$\pm$\scriptsize{0.020} & 0.869$\pm$\scriptsize{0.025} &0.771 & 0.488\\
    Mean-Pooling & 0.649 & 0.333 & 0.744$\pm$\scriptsize{0.030} & 0.454$\pm$\scriptsize{0.053} & 0.924$\pm$\scriptsize{0.020} & 0.849$\pm$\scriptsize{0.017} &0.772 & 0.545\\
    ABMIL~\cite{abmil} & \underline{0.699} & 0.333 & 0.797$\pm$\scriptsize{0.038} & \underline{0.487$\pm$\scriptsize{0.074}} & 0.944$\pm$\scriptsize{0.015} & \underline{0.867$\pm$\scriptsize{0.034}} &0.813 & 0.562\\
    CLAM-MB~\cite{CLAM} & 0.693 & 0.264 & 0.780$\pm$\scriptsize{0.038} & 0.469$\pm$\scriptsize{0.073} & 0.944$\pm$\scriptsize{0.018} & 0.864$\pm$\scriptsize{0.033} & 0.806 & 0.532\\
    DSMIL~\cite{dsmil} & 0.667 & 0.333 & 0.771$\pm$\scriptsize{0.037} & 0.478$\pm$\scriptsize{0.079} & 0.933$\pm$\scriptsize{0.020} & 0.860$\pm$\scriptsize{0.022} & 0.790 & 0.557\\
    DTFDMIL~\cite{dtfdmil} & 0.697 & \underline{0.368} & \underline{0.799$\pm$\scriptsize{0.039}} &0.486$\pm$\scriptsize{0.040} & \underline{0.945$\pm$\scriptsize{0.023}} & 0.839$\pm$\scriptsize{0.059} & \underline{0.814} & \underline{0.564}\\
    TransMIL~\cite{transmil} & 0.688 & 0.345 & 0.705$\pm$\scriptsize{0.028} & 0.328$\pm$\scriptsize{0.070} & 0.928$\pm$\scriptsize{0.021} & 0.848$\pm$\scriptsize{0.035} & 0.774 & 0.506\\
    S4MIL~\cite{S4MIL} & 0.676 & 0.299 & 0.776$\pm$\scriptsize{0.046} & 0.469$\pm$\scriptsize{0.062} & 0.935$\pm$\scriptsize{0.019} & 0.856$\pm$\scriptsize{0.027} & 0.796 & 0.541\\
    MambaMIL & \textbf{0.718} & \textbf{0.379} & \textbf{0.803$\pm$\scriptsize{0.040}} & \textbf{0.498$\pm$\scriptsize{0.073}} & \textbf{0.947$\pm$0.020} & \textbf{0.870$\pm$\scriptsize{0.037}} & \textbf{0.822} & \textbf{0.582}\\

    \bottomrule
    \end{tabular}}
    \label{tab:subtyping}
\end{table*}

\subsection{Comparison Results}
\noindent\textbf{Survival Prediction}. As presented in Table~\ref{tab:survival}, we conduct comparison experiments with two distinct feature settings on seven TCGA cancer datasets. The results demonstrate that MambaMIL achieves the best performance on all benchmarks compared to the state-of-the-art methods. Under the two feature sets, MambaMIL outperforms the second-best performance method by 2.6\% and 2.7\% on mean performance across all seven datasets. 

\noindent\textbf{Cancer Subtyping}. Table~\ref{tab:subtyping} shows experimental results on two datasets, encompassing both binary and multiple classification tasks. Compared to the state-of-the-art, our proposed MambaMIL demonstrates outstanding performance, attaining an AUC of 80.4\% on the BRACS dataset and 95.9\% on the NSCLC dataset. Notably, MambaMIL employs the same aggregation module as ABMIL but significantly outperforms it, with significant improvements of 3.9\% and 2.1\% in terms of AUC for BRACS and NSCLC datasets, respectively.

\begin{table*}[t]
	\centering
	\caption{Performance comparisons with different variations of Mamba.}
    \resizebox{11.2cm}{!}{
	\begin{tabular}	{ c  c | c c c  c c c c| c}
	\toprule
    \diagbox{\textbf{Method}}{\textbf{Dataset}} & 
     & \textbf{BLCA} & \textbf{BRCA} & \textbf{COADREAD} & \textbf{KIRC} & \textbf{KIRP} & \textbf{LUAD} & \textbf{STAD} & \textbf{MEAN}\\ 
         \midrule
    \rowcolor{mygray}
    \multicolumn{10}{c}{\textbf{\textit{ResNet-50}}}\\
    \midrule
Mamba && 0.622$\pm$\scriptsize{0.053} & 0.664$\pm$\scriptsize{0.034} & 0.650$\pm$\scriptsize{0.066} & 0.700$\pm$\scriptsize{0.058} & 0.734$\pm$\scriptsize{0.062} & 0.643$\pm$\scriptsize{0.027} & 0.621$\pm$\scriptsize{0.056} & 0.662\\
Bi-Mamba && 0.647$\pm$\scriptsize{0.024} & \textbf{0.675$\pm$\scriptsize{0.065}} & 0.662$\pm$\scriptsize{0.058} & 0.690$\pm$\scriptsize{0.048} & 0.737$\pm$\scriptsize{0.052} & 0.628$\pm$\scriptsize{0.059} & 0.622$\pm$\scriptsize{0.068} & 0.665\\
SR-Mamba && \textbf{0.652$\pm$\scriptsize{0.028}} & 0.673$\pm$\scriptsize{0.063} & \textbf{0.671$\pm$\scriptsize{0.066}} & \textbf{0.721$\pm$\scriptsize{0.064}} & \textbf{0.748$\pm$\scriptsize{0.094}} & \textbf{0.653$\pm$\scriptsize{0.059}} & \textbf{0.639$\pm$\scriptsize{0.076}} & \textbf{0.680}\\
    \bottomrule
	\end{tabular}}
	\label{tab:comparion}
\end{table*}

\begin{figure}[t]
    \centering
    \includegraphics[width=1.0\linewidth]{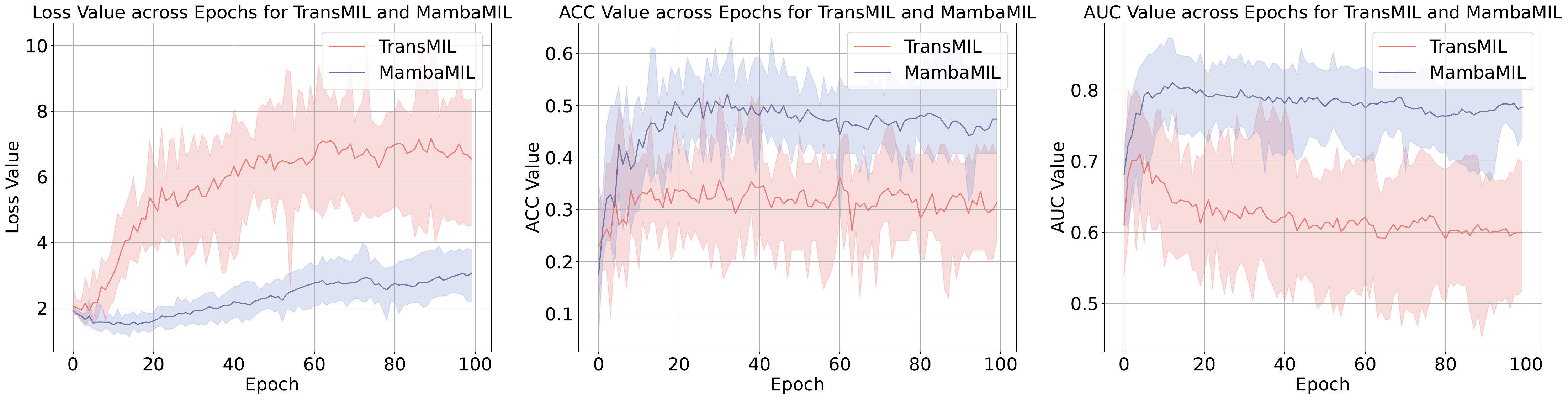}
    \caption{The performance comparison between TransMIL and our MambaMIL on the BRCAS validation set throughout the training process. }
    \label{fig:overfit}
\end{figure}

\subsection{Ablation Study}
To assess the effectiveness of SR-Mamba, we conduct extensive experiments to compare the performance of different variations of Mamba block: the vanilla Mamba~\cite{mamba}, Bidirectional Mamba (Bi-Mamba)~\cite{bimamba} and our proposed SR-Mamba, on survival prediction datasets. For a fair comparison of each specific dataset, we utilize the same setting to train these variants. As shown in Table~\ref{tab:comparion}, SR-Mamba surpasses the performance of Mamba and Bi-Mamba, which demonstrates the effectiveness of sequence reordering. Meanwhile, overfitting poses a substantial challenge in applying MIL methods for WSI analysis, especially for transformer-based methods like TransMIL. As illustrated in Fig.~\ref{fig:overfit}, during the training process, TransMIL displays clear signs of overfitting on the validation set, characterized by a significant increase in validation loss alongside decreases in both the ACC and the AUC metrics. In contrast, MambaMIL exhibits stable performance across the evaluation period, showcasing its strong ability to alleviate overfitting. This capability originates from the more discriminative representations extracted from various sequence orderings, akin to the effects of data augmentation, which significantly enhances the robustness of our proposed model.

\section{Conclusion}
In this paper, we introduce a novel Mamba-based MIL method, termed as MambaMIL, to tackle the challenges associated with long sequence modeling and overfitting, marking the first application of the Mamba framework in computational pathology. Our approach, based on the specially designed Sequence Reordering Mamba module (SR-Mamba), enables the effective leveraging of intrinsic global information contained within the long sequences of instances. The experimental results on nine benchmarks demonstrate that MambaMIL benefits from long sequence modeling and outperforms existing competitors under all metrics on
all benchmarks. 
Given the excellent performance of MambaMIL, we anticipate its application can be extended to other modalities in computational pathology, including genomics, pathology reports, and clinical data. This expansion would enable the leveraging of multi-modal information for effective and accurate diagnosis, prognosis, and therapeutic-response prediction.

\bibstyle{splncs04}
\bibliography{ys}

\clearpage

\appendix
\section*{Appendix}
\begin{algorithm}
\caption{SR-Mamba Block Process}
\begin{algorithmic}
\label{alg:SR-Mamba}
\pseudocodefont 
\Require instance sequence $X_{l-1} : (B,M,D)$
\Ensure instance sequence $X_l : (B,M,D)$
\LineComment{$B$: batch size, $M$: instance number, $D$: dimension}
\State $X'_{l-1} : (B,M,D) \leftarrow \text{LayerNorm}(X_{l-1})$ \\
$z : (B,M,E) \leftarrow \text{Linear}^z(X'_{l-1})$
\LineComment{Original Sequence: os}
\State $x_{\text{os}} : (B,M,E) \leftarrow \text{Linear}^{x0}(X'_{l-1})$
\LineComment{Reordered Sequence: rs}
\State $x_{\text{rs}} : (B,M,E) \leftarrow \text{Reordering}(\text{Linear}^{x1}(X'_{l-1}))$
\LineComment{process sequences with distinct orderings}
\For{$o$ in $\{\text{os}, \text{rs}\}$}
    \State $x_o : (B,M,E) \leftarrow \text{SiLU}(\text{Conv1d}_o(x))$
    \State $B_o : (B,M,N) \leftarrow \text{Linear}^B(x_o)$,\quad$C_o : (B,M,N) \leftarrow \text{Linear}^C(x_o)$
    \State $\Delta_o : (B,M,E) \leftarrow \log(1 + \exp(\text{Linear}^A(x_o) + \text{Parameter}^A))$
    \State $A_o : (B,M,E,N) \leftarrow \Delta_o \otimes \text{Parameter}^A$,\quad$B_o : (B,M,E,N) \leftarrow \Delta_o \otimes B_o$
    \State $y_o : (B,M,E) \leftarrow \text{SSM}(A_o,B_o,C_o)(x_o)$
\EndFor
\State $y_{\text{os}} : (B,M,E) \leftarrow y_{\text{os}} \odot \text{SiLU}(z)$,\quad $y_{\text{rs}} : (B,M,E) \leftarrow y_{\text{rs}} \odot \text{SiLU}(z)$
\LineComment{residual connection}
\State $X_l : (B,M,D) \leftarrow \text{Linear}(y_{\text{os}} + y_{\text{rs}}) + X_{l-1}$
\State \Return $X_l$
\end{algorithmic}
\label{algo}
\end{algorithm}

\begin{table*}[!htbp]
	\centering
\caption{Hyper-parameter configurations.}
	\resizebox{12.2cm}{!}{
	\begin{tabular}{c|c c c c c c c c c}
	\toprule
    \textbf{\textit{Datasets}} & \textbf{BLCA} & \textbf{BRCA} & \textbf{COADREAD} & \textbf{KIRC} & \textbf{KIRP} & \textbf{LUAD} & \textbf{STAD}& \textbf{BRACS}&  \textbf{NSCLC}\\
    \midrule
     \textbf{Sample} & 437  & 1023 & 572 & 498 & 261 & 455 & 363& 545 & 1053\\
    \textbf{Min Length} & 414  & 283 & 30 & 319 & 383 & 85 & 26 & 49& 85\\
    \textbf{Max Length} & 34174  & 36618 & 27418 & 30679 & 62235 & 45785 & 26130 & 26600 & 45785\\
    \textbf{Average Length} & 14419  & 8893 & 6953 & 12121 & 12424 & 9897 & 9402 & 7812& 10515\\
    \textbf{Learning Rate} & 2e-4  & 2e-5 & 2e-5 & 2e-4 & 2e-4 & 2e-4 & 2e-4 & 1e-5 & 2e-5\\
    \textbf{Segment Size $R$} & 5 & 5 & 10 & 10 & 10 & 5 & 5 & 10 & 5\\
    
    \bottomrule
	\end{tabular}}
	\label{tab:hyper-parameter}
\end{table*}

\end{document}